%% file: iccvw23.tex
\documentclass[10pt,twocolumn,letterpaper]{article}

\usepackage{iccv}
\usepackage{times}
\usepackage{epsfig}
\usepackage{graphicx}
\usepackage{amsmath}
\usepackage{amssymb}

\usepackage[pagebackref=true,breaklinks=true,letterpaper=true,colorlinks,bookmarks=false]{hyperref}

\usepackage{bbm}  % indicator function
\usepackage{booktabs} % toprule, cmidrule will affect the column line
\usepackage{multirow} % table multirow
\usepackage{xspace} % define a white space
\usepackage{enumitem}
\usepackage{footnote}

\usepackage{amsfonts}  % math font
\usepackage{xpatch}  % beamer ppt text align
\usepackage{xcolor,colortbl}
\usepackage{array, boldline, makecell} % makecell: table column width
\usepackage{arydshln}  % dot line in table
\usepackage{soul} %\so{1.letterspacing} \ul{2.underlining} \st{3.striking out} \hl{4.highlighting} \caps{5.CAPITALS, Small Capitals}
\usepackage{stackengine} % eq align

\usepackage{pifont}% 

\usepackage{mathtools}
\usepackage{lipsum}
\usepackage{marvosym}

\iccvfinalcopy % *** Uncomment this line for the final submission

\def\haiming{\textcolor{black}}

\ificcvfinal\fi

\begin{document}

%%%%%%%%% TITLE
\title{Progressive Feature Adjustment for \\ Semi-supervised Learning from Pretrained Models}

\author{Hai-Ming Xu$^{1}$
\quad
Lingqiao Liu\textsuperscript{1\Letter}
\quad
Hao Chen$^{2}$
\quad
Ehsan Abbasnejad$^{1}$
\quad
Rafael Felix$^{1}$\\
$^{1}$Australian Institute for Machine Learning, University of Adelaide
\quad
$^{2}$Zhejiang University\\
{\tt\small \{hai-ming.xu, lingqiao.liu, ehsan.abbasnejad, rafael.felixalves\}@adelaide.edu.au}\\
{\tt\small haochen.cad@zju.edu.cn}
}

\maketitle
% Remove page # from the first page of camera-ready.
\ificcvfinal\thispagestyle{empty}\fi

%%%%%%%%% ABSTRACT
\begin{abstract}
   As an effective way to alleviate the burden of data annotation, semi-supervised learning (SSL) provides an attractive solution due to its ability to leverage both labeled and unlabeled data to build a predictive model. While significant progress has been made recently, SSL algorithms are often evaluated and developed under the assumption that the network is randomly initialized. This is in sharp contrast to most vision recognition systems that are built from fine-tuning a pretrained network for better performance. While the marriage of SSL and a pretrained model seems to be straightforward, recent literature suggests that naively applying state-of-the-art SSL with a pretrained model fails to unleash the full potential of training data. In this paper, we postulate the underlying reason is that the pretrained feature representation could bring a bias inherited from the source data, and the bias tends to be magnified through the self-training process in a typical SSL algorithm. To overcome this issue, we propose to use pseudo-labels from the unlabelled data to update the feature extractor that is less sensitive to incorrect labels and only allow the classifier to be trained from the labeled data. More specifically, we progressively adjust the feature extractor to ensure its induced feature distribution maintains a good class separability even under strong input perturbation. Through extensive experimental studies, we show that the proposed approach achieves superior performance over existing solutions.
\end{abstract}

%%%%%%%%% BODY TEXT
\section{Introduction}
\label{sec:intro}
Semi-supervised learning (SSL) is considered one of the most practical learning paradigms which can leverage both labeled and unlabeled samples to build a prediction model~\cite{chapelle2009semi}. With the rapid development of deep neural networks (DNNs), extensive research on deep SSL methods~\cite{laine2016temporal,Tarvainen17MeanTeacher,miyato2018virtual,verma2021interpolation,berthelot2019mixmatch,xie2019unsupervised,sohn2020fixmatch} have been studied. Among those studies, most of them are evaluated and developed based on randomly initialized parameters. In recent years, the release and re-use of pretrained DNNs to alleviate training costs are becoming common practice for computer vision research and applications~\cite{zamir2018taskonomy}.
It seems that applying the existing SSL method to a pretrained model is straightforward. However, recent literatures~ \cite{zhou2018semi,wang2021self} suggest that such a naive solution fails to unleash the full potential of training data and there seems to be a big room to improve the performance of SSL when a pretrained model is used. 

\input{figure/fig_1st_impression}

In this study, we postulate the key issue preventing existing SSL solutions from attaining their full potential with pretrained models is due to the bias of the pretrained feature extractor: trained from the source domain data, e.g., ImageNet, the feature extractor may not be optimal for the target problem, e.g., fine-grained visual recognition. Such a bias could be a much more severe issue for SSL than for supervised learning. This is because the self-training (or pseudo-labeling) procedure commonly used in SSL tends to magnify the bias. For example, at the beginning of the training, 
a biased feature extractor and few labeled data could make the classifier vulnerable to spurious correlated patterns, resulting in a wrong prediction on the unlabeled data. The wrong prediction, however, will be fed back to the classifier again as pseudo labels and reinforce the bias. While SSL from randomly initialized network also suffer from the incorrect pseudo-labels, their feature extractors do not have the bias inherited from the source domain and thus could be easier adjusted through standard SSL methods towards the target problem. In reality, the bias of the feature extractor leads to the phenomenon that the SSL process from a pretrained model is less likely to correct its wrong prediction at the early training stage, as shown in Figure~\ref{fig:1st-impression}. 

In this work, we find that a surprisingly simple solution can largely resolve such an issue: we do not use unlabeled data and the corresponding pseudo-label to update the classifier but the feature extractor. Only labeled data are used to train the classifier. The rationale for this strategy is that the feature extractor is more tolerant to the label noise since it does not directly produce the final prediction, and with a good feature representation, it is possible to achieve good performance with only a few labeled data. More specifically, we require the feature representation should become closer to its class center while being pushed away from other class centers. Inspired by FixMatch~\cite{sohn2020fixmatch}, we also expect the above property holds for strongly-augmented data. The proposed method only modifies existing FixMatch by changing a few lines of code but has demonstrated a dramatic performance boost and even outperforms other carefully designed \haiming{contrastive-learning-based approaches} by a large margin. In summary, the main contribution of this paper are as follows:
\begin{itemize}[itemsep=2pt,topsep=0pt,parsep=0pt]
\item We provide an insight into the issue that standard SSL methods perform unsatisfactory with pretrained models and provide empirical evidence for a better understanding.

\item We discover that a simple solution that can significantly improve SSL from a pretrained model. Note that we do not claim the operation of aligning feature to its class-wise embedding is our novelty but the discovery that such a simple strategy can be a good solution to our studied issue.

\item We establish a strong baseline for SSL with pretrained model, providing practioner a simple-to-use solution for practical semi-supervised classification. 
\end{itemize}

%-------------------------------------------------------------------------
\section{Related Work}
\label{sec:related_work}
Semi-supervised learning (SSL) has experienced rapid progress with the development of deep neural networks (DNNs)~\cite{berthelot2019mixmatch,berthelot2019remixmatch,sohn2020fixmatch,zhang2021flexmatch,usb2022}. The current state-of-the-art SSL approaches~\cite{laine2016temporal,Tarvainen17MeanTeacher,miyato2018virtual,berthelot2019mixmatch,sohn2020fixmatch,Lee13Pseudolabel,xu2022semi} usually depend on the consistency regularization~\cite{berthelot2019mixmatch} and pseudo-labeling~\cite{Lee13Pseudolabel}. A common framework is to employ two processes: one process generates a prediction target, usually in the form of pseudo-labeling~\cite{sohn2020fixmatch,Lee13Pseudolabel}, but could also be logits~\cite{Tarvainen17MeanTeacher} or other supervision forms~\cite{berthelot2019mixmatch,xu2021semi,xu-etal-2022-progressive}. Then the generated pseudo supervision will be used to update the network with a different input, e.g., a different augmentation of the original input image~\cite{sohn2020fixmatch,clark2018semi}, mixed image~\cite{berthelot2019mixmatch}, or a network with different parameters~\cite{Tarvainen17MeanTeacher}. 

However, existing SSL approaches are primarily optimized from randomly initialized weights, and recent works~\cite{zhou2018semi,wang2021self} show that the impressive performance improvement of these standard SSL methods (includes the state-of-the-art method FixMatch~\cite{sohn2020fixmatch}) will disappear when models are training from a pretrained model. 
\haiming{Despite the initial findings reported in~\cite{zhou2018semi}, it still lacks a clear picture of why existing SSL methods perform unsatisfactory when pretrained models are used. In this paper, we investigate the optimization procedure of SSL from pretrained models and provide some empirical evidences to reveal barriers that limit performance. Based on the analysis, we further propose a feature adjustment module to progressively adjust the feature extractor and achieve great performance improvement on multiple vision benchmarks.}

\section{Preliminary of Semi-supervised Learning}
\label{subsec:bg}
In semi-supervised learning, two sets of samples are normally provided: $\{x^1_l, x^2_l, \cdots, x^{N_l}_l \} \in \mathcal{X_L}$ whose annotations $\{y^1_l, y^2_l, \cdots, y^{N_l}_l \}\in \mathcal{Y_L}$ are available and $\left \{x^1_u, x^2_u, \cdots, x^{N_u}_u\right \} \in \mathcal{X_U}$ where $N_u \gg N_l$ but without accessing label information. 

Although there are many existing SSL methods in the literature, this work mainly takes one of the state-of-the-art approaches FixMatch~\cite{sohn2020fixmatch} as an example. It is because FixMatch successfully integrates two popular techniques in SSL together, i.e., consistency regularization and pseudo-labeling, through decoupling the artificial label generation and model update with weak and strong data augmentations. Specifically, for samples from $\mathcal{X_L}$, the model $\mathcal{M}$ is trained via a standard classification loss. For each unlabeled data $x^i_u \in \mathcal{X_U}$, ``hard" pseudo label is firstly produced on the weakly augmented image
\begin{align}
    p(y|x_u^i)  = \mathcal{M} \left (A_0(x^i_u)\right ); \ \tilde{y} = \mathop{\text{argmax}}_{c} p(y=c | x^i_u), %\nonumber
\end{align}
where $A_0(\cdot)$ denotes weak data augmentation. Then, the model is optimized to have a consistent prediction on the strongly augmented image
\begin{equation}
    \mathcal{L}_u = \frac{1}{|\mathcal{B}_u|} \sum_{i\in \mathcal{B}_u} \mathbbm{1}\bigl( p(\tilde{y} | x^i_u) \geq \tau \bigr ) \text{CE} \Bigl(\mathcal{M}\bigl (A_1(x^i_u)\bigr ), \ \tilde{y} \Bigr ), %\nonumber
\end{equation}
where $p(\tilde{y} | x^i_u)$ means the $\tilde{y}$-th output probability on weakly augmented $x^i_u$. Here, $\mathbbm{1}(\cdot)$ is the indicator function to select samples whose predicting confidence is greater than a pre-defined confidence threshold $\tau$. Additionally, $\text{CE}(\cdot, \cdot)$ is the standard cross-entropy loss, $A_1(\cdot)$ is the strong augmentation and $\mathcal{B}_u$ denotes the size of unlabeled samples in one mini-batch.

\section{Semi-supervised Learning from Pretrained Models}
In this section, we first investigate the bias inherented in pretrained models and how existing SSL methods are troubled to achieve satisfactory performance. Then, we propose a feature adjustment module for SSL to resist the bias.

\subsection{Pretrained Models as a Double-edged Sword for SSL}
\label{sec:empirical_study}
With the growth of data and the enrichment of computing resources, developing powerful pretrained models have attracted increasing attention from both the academia and industry~\cite{devlin2018bert,he2019momentum,chen2020simple,radford2021learning,yuan2021florence,zeng2021pangu}. On the one hand, pre-training on large-scale images gives models the general feature extraction ability for various kinds of downstream applications~\cite{zhou2022learning,matsoukas2022makes,yang2022unified}. On the other hand, the generated feature representations will inevitably bring a bias inherited from the source data, and thus it is usually necessary to fine-tune the model on target datasets for effective usage of pretrained models~\cite{wang2017growing,guo2019spottune,howard2018universal}. 

For the SSL scenario, it is natural to expect that state-of-the-art performance can be achieved by applying the state-of-the-art SSL method to a pretrained model, i.e., semi-supervised fine-tuning. However, evidence from recent literatures~\cite{zhou2018semi,wang2021self} show that this solution is far from the best, and there is a big room to improve SSL when a pretrained model is used. This motivates us to revisit SSL and understand what hinders it from achieving its full potential. We postulate the major issue is that the use of pretrained model is a double-edged sword for SSL: on the one hand, it brings the prior knowledge learned from the source data and boosts the performance. On the other hand, 
it also introduces a strong prediction bias inherited from the source data. After all, the feature learned from the source domain data may not be optimal for the target task. In effect, the prediction bias will encourage the classifier to use certain features or visual patterns for prediction, especially when the labeled data is limited in quantity and diversity. 
However, not all those visual patterns are true causal factors to determine the class and the spurious correlation might be mistakenly identified during training, e.g., prediction could rely on the clue from background~\cite{shu2022improving}.

Worse, as most of the state-of-the-art SSL methods~\cite{sohn2020fixmatch,berthelot2019mixmatch,Lee13Pseudolabel,berthelot2019remixmatch,zhang2021flexmatch} are built upon self-training, a.k.a., pseudo-labeling framework, which generates pseudo labels from the prediction on the unlabeled data, such a process tends to further magnify the prediction bias. For example, we can consider the scenario of applying FixMatch, one of the most commonly used SSL methods, with a pretrained model. At the beginning of training, due to the limited amount of labeled (and pseudo-labeled) samples and biased feature representation, the learned classifier tends to be affected by the spurious correlation between features and class labels. Then if the classifier generates incorrect pseudo labels from unlabeled data, the bias will thus reinforce itself by further training with such pseudo labels. Consequently, this will make the SSL less prone to correct its wrong prediction made during the training process.

In order to verify our assumption, we conduct an empirical analysis on FixMatch with pretrained models. Specifically, we introduce two measurements called correct-keeping rate (CKR) and error-correcting rate(ECR). The former is defined as a percentage of the initially\footnote{``Initially'' here means an unlabeled sample was falsely labeled for the first time in the whole training process.} correctly labeled data keeping their correct class at a given iteration, while the latter one is defined as the percentage of the initially incorrectly labeled data being predicted to the correct class at a given iteration. The statistical results are shown in Figure~\ref{fig:1st-impression}. Then we can observe that FixMatch (with a pretrained model) has a descent CKR metric when training converges. However, the ECR of FixMatch (with a pretrained model) quickly reaches a plateau and has a poor ECR metric. This issue becomes more evident by comparing its ECR with the proposed method. 

\input{figure/fig_main_structure}

\subsection{Progressive Feature Adjustment}

The above analysis suggests that the standard SSL approach could suffer more from the biased feature extractor and we may need a special process to alleviate the impact of bias. In this work, we propose to only use the labeled data to train the classifier, while pseudo-labels generated from the larger amount of unlabeled samples will only be used to update the feature extractor. In this way, unlabeled data influences the classifier indirectly by producing better feature representations. As the classifier is always trained on noise-free labeled data, even if the feature representation is imperfect, the classifier can suppress the noisy dimensions and identify the discriminative patterns in the feature representation. Thus the feature extractor can be more tolerant to the noise in pseudo-labels. It seems that one drawback of the above method is the lack of training examples for the classifier. However, since a pretrained feature extractor has already been able to provide a reasonable starting point and will be further refined by the proposed progressive adjustment method, training the classifier on a limited number of samples can guarantee a good performance.
The overall architecture is shown in Figure~\ref{fig:main_structure}. 

Specifically, given a batch of labeled samples $(x_l, y_l)\in \mathcal{B}_l$, both of the feature extractor $f$ and the linear classifier $\mathbf{W}:=\{\mathbf{w}_1, \mathbf{w}_2, \cdots, \mathbf{w}_c, \cdots \}$ will be optimized together
\begin{equation}\label{eq:lce}
\begin{aligned}
    p(y = c|x^i_l) & = \frac{\exp \bigl (\mathbf{w}_c^T f(x_l^i)\bigr)}{\sum_j \exp\bigl(\mathbf{w}_j^T f(x_l^i)\bigr)}, \nonumber
\end{aligned}
\end{equation}
\begin{equation}
    \mathcal{L}_l = \frac{1}{|\mathcal{B}_l|} \sum_{i\in \mathcal{B}_l} \text{CE}\bigl (p(y|x_l^i),y^i_l\bigr)
\end{equation}where $\mathbf{w}_c$ denotes the classifier for class $c$. $\text{CE}(\cdot, \cdot)$ is the standard cross-entropy loss.

For a batch of unlabeled samples $x_u\in\mathcal{B}_u$, we utilize the up-to-date classifier to generate posterior probability estimation $p(y|x^i_u)$
\begin{equation}\label{eq:pl}
\begin{aligned}
    p(y = c|x^i_u) & = \frac{\exp \Bigl (\mathbf{w}_c^T f\bigl (A_0(x_u^i)\bigr)\Bigr)}{\sum_j \exp\Bigl(\mathbf{w}_j^T f\bigl (A_0(x_u^i)\bigr)\Bigr)},
\end{aligned}
\end{equation}where $f(A_0(x_u^i))$ denotes the feature extracted by first going through the weak data augmentation module $A_0$ and then the feature extractor $f$ (same as that in FixMatch). The class corresponding to the maximal posterior probability is the predicted class of the given unlabeled sample, that is, $\tilde{y}^i = \mathop{argmax}_{c} \ p(y = c|x^i_u)$. If $p\bigl (\tilde{y}^i | x^i_u\bigr) \geq \tau $ where $\tau$ is the confidence threshold which can be a fixed scalar, e.g., 0.95 as in FixMatch~\cite{sohn2020fixmatch} or a dynamic generated scalar as in FlexMatch~\cite{zhang2021flexmatch}, then $\tilde{y}^i$ will be used as a pseudo-label for the corresponding unlabeled sample.

Instead of using the pseudo-labeled unlabeled samples to train on the feature extractor and classifier altogether, as in FixMatch, we propose to use them to adjust the feature extractor only. Without introducing an additional linear connected layer, we maintain a set of class-wise embeddings $\{\mu_c\}_0^C$ and try to minimize the following loss:
\begin{small}
\begin{align}
\label{eq:u_proto}
    &\hat{\mathcal{L}}_f = \frac{1}{|\mathcal{B}_u |} \sum_{i\in \mathcal{B}_u} \mathbbm{1} \bigl (p(\tilde{y}^i|x^i_u) \geq \tau \bigr) \mathcal{L}_f(x^i_u,\tilde{y}^i) + \frac{1}{|\mathcal{B}_l|} \sum_{i\in \mathcal{B}_l}\mathcal{L}_f(x^i_l,y^i_l) . \nonumber \\
       & where, \mathcal{L}_f(x,y) = -\log \frac{\exp\biggr(\cos\Bigl(\mathbf{\mu}_{y}, f\bigl (A_1(x)\bigr )\Bigr)/T\biggr)}{\sum_j \exp\biggl(\cos\Bigl(\mathbf{\mu}_j, f\bigl (A_1(x)\bigr )\Bigr )/T\biggr)},
\end{align}
\end{small}where $cos(\cdot, \cdot)$ denotes the cosine similarity and $T$ is a temperature hyperparameter. We empirically set $T=0.1$ in our study. $A_1$ denotes a different type of data augmentation to $A_0$ and we use RandAugment~\cite{cubuk2020randaugment} followed by Cutout~\cite{devries2017improved} as the strong data augmentation $A_1$. $\mathbf{\mu}_c$ is the running class mean vector for the $c$-th class. 

In effect, the above loss function will pull features from the same class closer while push features from different classes far apart. For labeled data, we could use the ground-truth class label for assigning samples to their corresponding class embeddings. For unlabeled data, we use pseudo-labels instead and only apply the loss to samples that can generate pseudo-labels. Also, motivated by FixMatch, we propose to apply this loss on strongly augmented data to further avoid the confirmation bias. Note that the above loss also implicitly encourages same-class features from the labeled and unlabeled data move closer relative to the distance to other class samples. Thus it tends to make the classifier learned from the labeled data more generalizable to unlabeled data.

To sum up, we train the classifier with labeled data only and both labeled and unlabeled data with the loss in Eq.~\ref{eq:u_proto}. The overall loss function $\mathcal{L}$ is the weighted summation of both: $\mathcal{L} = \mathcal{L}_l + \lambda \cdot \hat{\mathcal{L}}_f$, where $\lambda$ is the fixed weight hyperparameter.

In order to adjust the feature extractor efficiently, oracle class-wise embeddings of target dataset should be an ideal choice. However, it is unrealistic due to the lack of annotations for unlabeled samples and the inherent bias in the pretrained feature extractor. Thus, we propose to progressively update these class-wise embeddings from both labeled and unlabeled data.
For labeled data, $\mathbf{\mu}_c$ is updated via
\begin{align}\label{eq:mu_l}
    \mathbf{\mu}^{new}_c =  \beta \mathbf{\mu}^{old}_c + (1-\beta) f\bigl(A_1(x_l^i)\bigr) \mathbbm{1}(y^i_l = c), %\nonumber
\end{align}
and for unlabeled data, $\mathbf{\mu}_c$ is updated via
\begin{align}\label{eq:mu_u}
    \mathbf{\mu}^{new}_c =  \beta \mathbf{\mu}^{old}_c + (1-\beta) f\bigl (A_1(x_u^i)\bigr) \mathbbm{1}\Bigl( p\bigl (c | A_0(x^i_u)\bigr) \geq \tau \Bigr ), %\nonumber
\end{align}where the indicator function $\mathbbm{1}(y^i_l = c)$ selects samples from the $c$-th class from the labeled data, the indicator function $\mathbbm{1}\Bigl( p^t\bigl (c | A_0(x^i_u)\bigr) \geq \tau \Bigr )$ selects unlabeled samples that are confidently classified into the $c$-th class by the classifier. This is identical to the criterion of generating pseudo-labels. $\beta$ is a momentum term that controls how far the class-wise feature reaches into embedding history.

\section{Experimental results}
\label{sec:exp}
In this section, we compare our approach with several SSL methods with pretrained models.

\subsection{Experimental details}
We strictly follow \cite{wang2021self} to design our experiment, including the evaluation datasets and pretrained model choices. We made such a choice since Self-Tuning~\cite{wang2021self} has demonstrated the state-of-the-art performance and its experimental evaluation is comprehensive and realistic. 
Some experimental details are as follows:

\smallskip
\noindent \textbf{Datasets}: Following the protocol of \cite{wang2021self}, which addressed the same research problem as this paper, four vision benchmarks are evaluated, i.e., \textit{FGVC Aircraft}~\cite{MajiRKBV13Air}, \textit{Stanford Cars}~\cite{KrauseStarkDengFei-Fei_3DRR2013}, \textit{CUB-200-2011}~\cite{WahCUB_200_2011}, and \textit{CIFAR-100}~\cite{Krizhevsky09cifar}. Specifically, the first three are challenging fine-grained classification datasets and label proportions ranging from 15\% to 50\% are tested. Label partition of \textit{CIFAR-100} follows the standard SSL protocol: 4/25/100 labeled images per class. 

\smallskip
\noindent \textbf{Methods}: Nine popular deep SSL approaches are included for comparison, i.e., {$\Pi$-model}~\cite{Laine17Pimodel}, {Pseudo-Labeling}~\cite{Lee13Pseudolabel},  {Mean Teacher}~\cite{Tarvainen17MeanTeacher}, {UDA}~\cite{xie2019unsupervised}, {FixMatch}~\cite{sohn2020fixmatch}, {FlexMatch}~\cite{zhang2021flexmatch}, {SimCLRv2}~\cite{chen2020big}, {FixMatch+AKC+ARC}~\cite{abuduweili2021adaptive} and {Self-Tuning}~\cite{wang2021self}. Meanwhile, performance of Fine-Tuning on labeled data is also reported for a reference baseline. 
For our approach, we also consider a simple extension by incorporating it into the recently proposed consistency-based SSL method FlexMatch~\cite{zhang2021flexmatch}, which uses dynamically assigned threshold with a FixMatch framework. We call this extension \textbf{Ours+}. 
Note that it shows our approach can still boost the performance even with more advanced SSL algorithms. In our work, all experiments were implemented in PyTorch and run on a GeForce RTX 2080Ti GPU with 11GB memory.

\smallskip
\noindent \textbf{Pretrained models}: Following \cite{wang2021self}, three models pretrained on ImageNet~\cite{deng2009imagenet} are chosen for evaluation, i.e., ResNet-50~\cite{he2016deep} and EfficientNet~\cite{Tan19EfficientNet} which are pretrained in a supervised way, and a ResNet-50 which is trained through an unsupervised learning method MoCo v2~\cite{he2019momentum}. 

\input{table/table_main_results_cifar.tex}
\input{table/table_main_results.tex}

\subsection{Train from Supervised Pretrained Models}
\label{subsec:sup_pretrained_model}
In this section, we compare various SSL methods trained from supervised pretrained models.

\smallskip
\noindent \textbf{Fine-grained Classification benchmarks}: 
We use a ResNet-50 network, which is supervised pretrained on ImageNet, to initialize all SSL models. The results are shown in Table~\ref{tab:main_res}. It is clear that our proposed method achieves overall significant improvement than other comparing SSL approaches. Specifically, compared with traditional SSL methods, our approach increases the test accuracy by a large margin on all kinds of partitions of three benchmarks. Taking the state-of-the-art method FixMatch~\cite{sohn2020fixmatch} as an example, the performance gain of our approach exceeds 10 percent on both \textit{Stanford Cars} and \textit{CUB-200-2011} with 15\% labels. 
This is thanks to the proposed feature adjustment module in our approach which greatly reduces the bias inherented in the pretrained model.
Furthermore, our approach is also superior to the recently proposed Self-Tuning method~\cite{wang2021self} especially when labels are limited, e.g., only 15\% training samples are labeled. 
When the CPL module proposed in FlexMatch~\cite{zhang2021flexmatch} is added to our approach, Ours+ leads to a further performance boost.

\smallskip
\noindent \textbf{Standard SSL benchmarks}: We choose \textit{CIFAR-100} dataset~\cite{Krizhevsky09cifar} which is one of the most challenging datasets among standard SSL benchmarks to evaluate SSL methods from a pretrained model. 
Due to the lack of open-resourced pretrained checkpoints on WideResNet-28-8 model~\cite{ZagoruykoK16WideResNet}, EfficientNet-B2 model~\cite{Tan19EfficientNet} supervised pretrained on ImageNet is adopted in this work. Table~\ref{tab:main_res_c100} presents the error rates of each method. Our proposed method yields the best performance among the comparing methods.

\subsection{Train from Unsupervised Pretrained Models}
Various semi-supervised learning approaches have been shown to benefit from supervised pretrained models in Section~\ref{subsec:sup_pretrained_model}, we continue to study the transfer effect from MoCov2~\cite{he2019momentum} which is pretrained on ImageNet without using any annotations. As the test accuracy presented in Figure~\ref{fig:main_res_moco_cub}, our best performed model, Ours+, excels to other semi-supervised learning baselines.

\input{figure/fig_tsne_feat}
\input{figure/fig_moco_cub}

\subsection{Ablation Study}
\label{sec:abl_study}
We are interested in ablating our approach from the following perspective views:

\subsubsection{The distribution of feature representation:} 
In our approach, the progressive feature adjustment module is introduced to update the feature extractor separately for alleviating the bias inherented in pretrained models. Therefore, we are interested in the effect of using such module or not on the feature distribution.
Figure~\ref{fig:tsne} presents the feature distribution of FixMatch and ours approach for some classes of \textit{FGVC Aircraft} with t-SNE~\cite{van2008visualizing}. We can find that our method encourages same class features to be close to each other while staying away from the other class samples and produces a more distinguishable distribution for the target data, while FixMatch suffers from the inherented bias of pretrained model and poorly adapts the feature distribution given the observation whose features from different classes are mixed together, thus its performance is heavily limited.

\input{table/table_supp_random_init.tex}

\subsubsection{Is our approach effective for randomly initialized network?}
In our formulation, the progressive feature adjustment module can be seen as a special SSL method for SSL from a pretrained model. So we are interested to know its effectiveness for randomly initialized network. To investigate this, we train our approach with a randomly initialized WideResNet-28-8~\cite{ZagoruykoK16WideResNet} network on CIFAR-100. As the results shown in Table~\ref{tab:supp_random_init}, our approach does not produce significant improvement as what we have observed in the SSL with pretrained models task. We postulate that this is because the feature extractor of a randomly initialized network does not inherited the prediction bias from the source domain, and thus the original design in FixMatch algorithm has already been sufficient to adjust the feature extractor for the target problem.

\input{table/table_abl_model_bias}

\input{figure/fig_threshold_sensitive}

\subsubsection{Does our approach work for other SSL method?}
We are also interested in if the proposed progressive feature adjustment module can be extended to other SSL methods. To investigate this, we apply this module to UDA~\cite{xie2019unsupervised} which is another popular consistency-regularization based SSL method. We conduct experiments on \textit{FGVC Aircraft} dataset and present the results in Table~\ref{tab:abl_model_bias}. As seen, by incorporating the proposed progressive feature adjustment module, we can significantly improve UDA in the SSL from pretrained models setting. This suggests that the proposed progressive feature adjustment module could be used to upgrade various consistency-regularization based SSL methods when pretrained models are available.  

\subsubsection{The ways of updating class-wise embeddings} In our approach, the class-wise embedding, i.e., class mean vectors, are dynamically updated from features of strongly augmented labeled images and features of strongly augmented unlabeled samples whose pseudo-supervisions are confident enough. In this section, we investigate two alternative strategies: 1) accumulate features of weakly augmented images to update mean vectors, 2) estimate parameters from features of unlabeled samples without a confidence threshold. As the results shown in Table~\ref{tab:abl_proto}, both of these two alternatives will result in a slight performance drop to our method.

\subsubsection{The sensitivity analysis of hyperparameter selection in our approach:} There are two hyperparameters in our method: one is the momentum term beta (i.e., $\beta$) for updating the class-wise embedding $\mu$ in Eq.~\ref{eq:mu_l} and Eq.~\ref{eq:mu_u}, and the other one is the balance weight lambda (i.e., $\lambda$) for the overall loss. As shown in Figure~\ref{fig:abl_hyper_sensitivity}, Our method is robust to the selection of both $\beta$ and $\lambda$ hyperparameters.

\input{table/table_abl_proto}

\section{Conclusion}
Semi-supervised learning from pretrained models is an encouraging research direction, because it combines the advantages of the two learning paradigms to achieve more data-efficient learning. Given the observations in the literature show that existing semi-supervised learning algorithms do not produce a satisfactory performance boost compared to their training-from-scratch version, we investigate the learning procedure of semi-supervised learning from pretrained models and find that the bias inherented in the original pretrained models may be magnified along the semi-supervised training. Empirical evidences are also provided for a better understanding. Based upon the analysis, we propose a progressive feature adjustment module to decouple the process of pseudo-supervision generation and model update and thus alleviate the bias successfully. Extensive experimental results on four vision benchmarks verify the effectiveness of our proposed approach.

\noindent\textbf{Acknowledgement.} This work was done in Adelaide Intelligence Research (AIR) Lab and Hai-Ming Xu and Lingqiao Liu are supported by the Centre of Augmented Reasoning (CAR).

{\small
\bibliographystyle{ieee_fullname}
\bibliography{egbib}
}

\end{document}

%% file: figure/fig_1st_impression.tex
\begin{figure}[t]
\centering
\includegraphics[width=\linewidth]{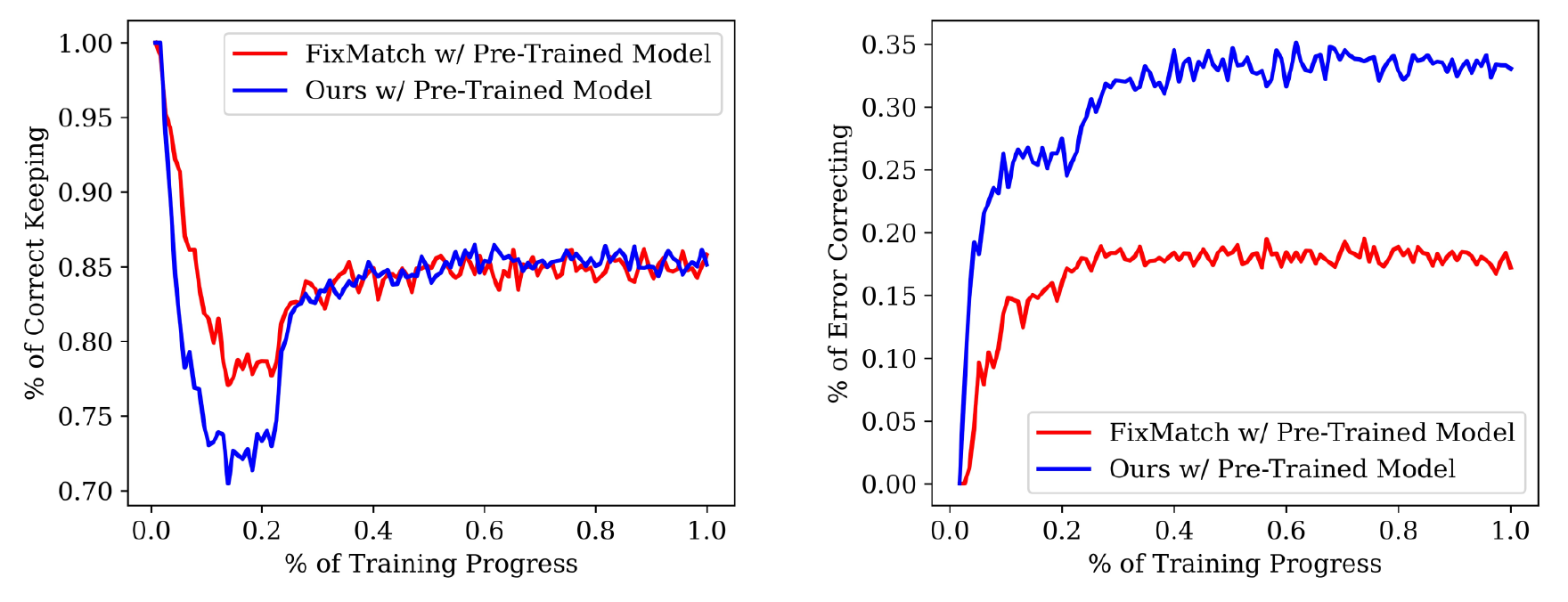}
\caption{Visualization of the correct keeping and error correcting ability of FixMatch and ours approach with a pretrained model. The y-axis of the left figure denotes the percentage of the initially correctly labeled data keeping their correct class at a given iteration. The y-axis of the right figure denotes the percentage of the initially incorrectly labeled data being predicted to the correct class at a given iteration. The experiment is conducted on \textit{FGVC Aircraft} dataset with 15\% labels. As seen, FixMatch with a pre-trained model shows weaker error correcting ability than the ours approach. Please refer to Section \ref{sec:empirical_study} for more details.}
\label{fig:1st-impression}
\end{figure}

%% file: figure/fig_main_structure.tex
\begin{figure*}[t]
\centering
\includegraphics[width=1.0\textwidth]{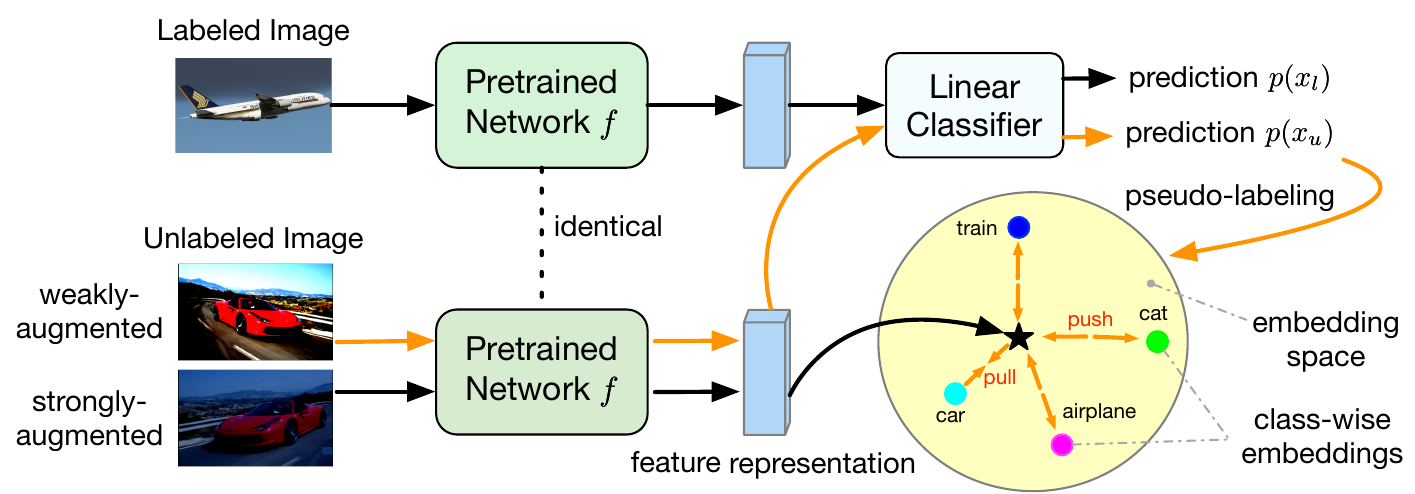}
\caption{Overview of our approach. The feature extractor is initialized with a pretrained model and the classifier is random initialized for the target dataset. In order to alleviate the bias as presented in Section~\ref{sec:empirical_study}, we let the classifier only trained on the labeled samples and use the large amount of unlabeled samples to adjust the feature extractor alone through pulling the immediate feature representation to its corresponding class embedding and pushing away to other class embeddings. The class-wise embeddings are progressively updated along the model optimization as presented in Eq.~\ref{eq:mu_l} and Eq.~\ref{eq:mu_u}.
}
\vspace{-0.5cm}
\label{fig:main_structure}
\end{figure*}

%% file: table/table_main_results_cifar.tex
\begin{table}[tbp]%\small
    \centering
    \begin{tabular}{l | c c c}
         \toprule
         \multirow{2}{*}{Method} & \multicolumn{3}{c}{Label Number} \\
         \cmidrule(lr){2-4}
         & 400 & 2500 & 10000 \\
         \hline
         Fine-Tuning (baseline) & 60.79 & 31.69 & 21.74 \\
         Pseudo-Labeling~\cite{Lee13Pseudolabel} & 59.21 & -- & -- \\
         MT~\cite{Tarvainen17MeanTeacher} & 60.68 & -- & -- \\
         UDA~\cite{xie2019unsupervised} & 58.32 & -- & -- \\
         FixMatch$\dag$~\cite{sohn2020fixmatch} & 52.88 & 25.63 & 18.38 \\
         FlexMatch$\dag$~\cite{zhang2021flexmatch} & 40.41 & 23.19 & 17.73 \\
         Self-Tuning~\cite{wang2021self} & 47.17 & 24.16 & 17.57 \\
         \textbf{Ours}$\dag$ & 45.48 & 23.12 & 16.89 \\
         \textbf{Ours+}$\dag$ & \textbf{37.36} & \textbf{22.06} & \textbf{16.58} \\
         \bottomrule
    \end{tabular}
    \caption{Error rates (\%) on \textit{CIFAR-100} with EfficientNet-B2. $\dag$ means ours implementation based on~\cite{wang2021self}.
    }
    \label{tab:main_res_c100}
    \vspace{-0.5cm}
\end{table}

%% file: table/table_main_results.tex
\begin{table*}[h!]
    \centering
    \begin{tabular}{c | l | c c c}
         \Xhline{1pt}
         \multirow{2}{*}{Dataset} & \multirow{2}{*}{Method} & \multicolumn{3}{c}{Label Proportion} \\
         \Xcline{3-5}{0.4pt}
          & & 15 \% & 30 \% & 50 \% \\
         \Xhline{0.6pt}
         \multirow{12}{*}{\textit{FGVC Aircraft}} 
         & Fine-Tuning (baseline) & 39.57{\scriptsize $\pm$0.20}& 57.46{\scriptsize $\pm$0.12}& 67.93{\scriptsize $\pm$0.28} \\
         & $\Pi$-model~\cite{Laine17Pimodel} & 37.32{\scriptsize $\pm$0.25} & 58.49{\scriptsize $\pm$0.26} & 65.63{\scriptsize $\pm$0.36} \\
		 & Pseudo-Labeling~\cite{Lee13Pseudolabel} &46.83{\scriptsize $\pm$0.30} &62.77{\scriptsize $\pm$0.31}&73.21{\scriptsize $\pm$0.39} \\
		 & Mean Teacher~\cite{Tarvainen17MeanTeacher} & 51.59{\scriptsize $\pm$0.23} & 71.62{\scriptsize $\pm$0.29} & 80.31{\scriptsize $\pm$0.32} \\
		 & UDA$\dag$~\cite{xie2019unsupervised} & 59.50{\scriptsize $\pm$0.36}  & 74.08{\scriptsize $\pm$0.41}  & 81.10{\scriptsize $\pm$0.42} \\
         & FixMatch$\dag$~\cite{sohn2020fixmatch} & 60.19{\scriptsize $\pm$0.43}& 75.28{\scriptsize $\pm$0.39} & 81.19{\scriptsize $\pm$0.41} \\
         & FlexMatch$\dag$~\cite{zhang2021flexmatch} &  63.21{\scriptsize $\pm$0.15} & 77.08{\scriptsize $\pm$0.34} & 82.56{\scriptsize $\pm$0.22} \\
         & SimCLRv2 ~\cite{chen2020big} & 40.78{\scriptsize $\pm$0.21}  & 59.03{\scriptsize $\pm$0.29}  & 68.54{\scriptsize $\pm$0.30} \\
         & FixMatch+AKC+ARC$\dag$~\cite{abuduweili2021adaptive} & 63.87{\scriptsize$\pm$0.41} & 75.99{\scriptsize$\pm$0.38} & 81.24{\scriptsize$\pm$0.31} \\
		 & Self-Tuning~\cite{wang2021self} & 64.11{\scriptsize $\pm$0.32} & 76.03{\scriptsize $\pm$0.25} & 81.22{\scriptsize $\pm$0.29}\\
         \Xcline{2-5}{0.4pt}
         & \textbf{Ours}$\dag$ & 69.64{\scriptsize $\pm$0.41} & 82.36{\scriptsize $\pm$0.44} & 85.02{\scriptsize $\pm$0.33} \\
		 & \textbf{Ours+}$\dag$ & \textbf{71.23}{\scriptsize $\pm$0.26} & \textbf{82.80}{\scriptsize $\pm$0.15} & \textbf{85.53}{\scriptsize$\pm$0.32} \\
        \Xhline{0.6pt}
		 
		 \multirow{12}{*}{\textit{Stanford Cars}} 
		 & Fine-Tuning (baseline) & 36.77{\scriptsize $\pm$0.12} & 60.63{\scriptsize $\pm$0.18} & 75.10{\scriptsize $\pm$0.21} \\
		 & $\Pi$-model~\cite{Laine17Pimodel} & 45.19{\scriptsize $\pm$0.21} & 57.29{\scriptsize $\pm$0.26} & 64.18{\scriptsize $\pm$0.29} \\
		 & Pseudo-Labeling~\cite{Lee13Pseudolabel} & 40.93{\scriptsize $\pm$0.23} &67.02{\scriptsize $\pm$0.19}& 78.71{\scriptsize $\pm$0.30} \\
		 & Mean Teacher~\cite{Tarvainen17MeanTeacher} & 54.28{\scriptsize $\pm$0.14} & 66.02{\scriptsize $\pm$0.21} & 74.24{\scriptsize $\pm$0.23} \\
		 & UDA$\dag$~\cite{xie2019unsupervised} & 61.88{\scriptsize $\pm$0.39} & 79.16{\scriptsize $\pm$0.36} & 86.79{\scriptsize $\pm$0.31} \\
		 & FixMatch$\dag$~\cite{sohn2020fixmatch} & 64.97{\scriptsize $\pm$0.37} & 81.23{\scriptsize $\pm$0.31} & 87.74{\scriptsize $\pm$0.35} \\
		 & FlexMatch$\dag$~\cite{zhang2021flexmatch} &  71.96{\scriptsize $\pm$0.28} & 83.81{\scriptsize $\pm$0.26} & 88.12{\scriptsize $\pm$0.21} \\
		 & SimCLRv2 ~\cite{chen2020big} & 45.74{\scriptsize $\pm$0.16} & 61.70{\scriptsize $\pm$0.18} & 77.49{\scriptsize $\pm$0.24} \\
         & FixMatch+AKC+ARC$\dag$~\cite{abuduweili2021adaptive} & 68.63{\scriptsize$\pm$0.38} & 82.81{\scriptsize$\pm$0.27} & 87.98{\scriptsize$\pm$0.32} \\
		 & Self-Tuning~\cite{wang2021self} & 72.50{\scriptsize $\pm$0.45} & 83.58{\scriptsize $\pm$0.28} & 88.11{\scriptsize $\pm$0.29} \\
        \Xcline{2-5}{0.4pt}
		 & \textbf{Ours}$\dag$ & 77.22{\scriptsize $\pm$0.42} & 86.91{\scriptsize $\pm$0.07} & 90.38{\scriptsize $\pm$0.16} \\
		 & \textbf{Ours+}$\dag$ & \textbf{79.70}{\scriptsize $\pm$0.31} & \textbf{87.92}{\scriptsize $\pm$0.32} & \textbf{90.71}{\scriptsize $\pm$0.13} \\
		 \Xhline{0.6pt}
		 
		 \multirow{12}{*}{\textit{CUB-200-2011}} 
		 & Fine-Tuning (baseline) & 45.25{\scriptsize $\pm$0.12} & 59.68{\scriptsize $\pm$0.21}  & 70.12{\scriptsize $\pm$0.29} \\
		 & $\Pi$-model~\cite{Laine17Pimodel} & 45.20{\scriptsize $\pm$0.23} & 56.20{\scriptsize $\pm$0.29} & 64.07{\scriptsize $\pm$0.32} \\
		 & Pseudo-Labeling~\cite{Lee13Pseudolabel} &45.33{\scriptsize $\pm$0.24}& 62.02{\scriptsize $\pm$0.31} & 72.30{\scriptsize $\pm$0.29} \\
		 & Mean Teacher~\cite{Tarvainen17MeanTeacher} &  53.26{\scriptsize $\pm$0.19} & 66.66{\scriptsize $\pm$0.20} & 74.37{\scriptsize $\pm$0.30} \\
		 & UDA$\dag$~\cite{xie2019unsupervised} & 52.23{\scriptsize $\pm$0.23}  & 67.93{\scriptsize $\pm$0.25}  & 75.63{\scriptsize $\pm$0.28} \\
		 & FixMatch$\dag$~\cite{sohn2020fixmatch} & 54.21{\scriptsize $\pm$0.26} & 69.28{\scriptsize $\pm$0.28} & 77.49{\scriptsize $\pm$0.31} \\
		 & FlexMatch$\dag$~\cite{zhang2021flexmatch} &  61.26{\scriptsize $\pm$0.18} & 71.62{\scriptsize $\pm$0.32} & 78.06{\scriptsize $\pm$0.21} \\
		 & SimCLRv2 ~\cite{chen2020big} & 45.74{\scriptsize $\pm$0.15} & 62.70{\scriptsize $\pm$0.24} & 71.01{\scriptsize $\pm$0.34} \\
         & FixMatch+AKC+ARC$\dag$~\cite{abuduweili2021adaptive} & 63.21{\scriptsize$\pm$0.35} & 73.61{\scriptsize$\pm$0.32} & 79.08{\scriptsize$\pm$0.29} \\
		 & Self-Tuning~\cite{wang2021self} & 64.17{\scriptsize $\pm$0.47} & 75.13{\scriptsize $\pm$0.35} & 80.22{\scriptsize $\pm$0.36} \\
         \Xcline{2-5}{0.4pt}
		 & \textbf{Ours}$\dag$ & 65.55{\scriptsize $\pm$0.21} & 74.99{\scriptsize $\pm$0.33} & 80.00{\scriptsize $\pm$0.11} \\
		 & \textbf{Ours+}$\dag$ & \textbf{68.06}{\scriptsize $\pm$0.22} & \textbf{76.09}{\scriptsize $\pm$0.34} & \textbf{80.40}{\scriptsize $\pm$0.21} \\
        \Xhline{1pt}
    \end{tabular}
    \caption{Test accuracy (\%) $\uparrow$ 
    on three fine-grained SSTL benchmarks. We empirically find strong augmentation for labeled data used in Self-Tuning~\cite{wang2021self} can bring performance gains to other SSL methods. Following the same setting of
    Self-Tuning, 
    Methods with $\dag$ are implemented by ourself based on the released codebase of Self-Tuning~\cite{wang2021self}. 
    }
    \label{tab:main_res}
    \vspace{-0.5cm}
\end{table*}

%% file: figure/fig_tsne_feat.tex
\begin{figure*}[t]
\centering
\includegraphics[width=\linewidth]{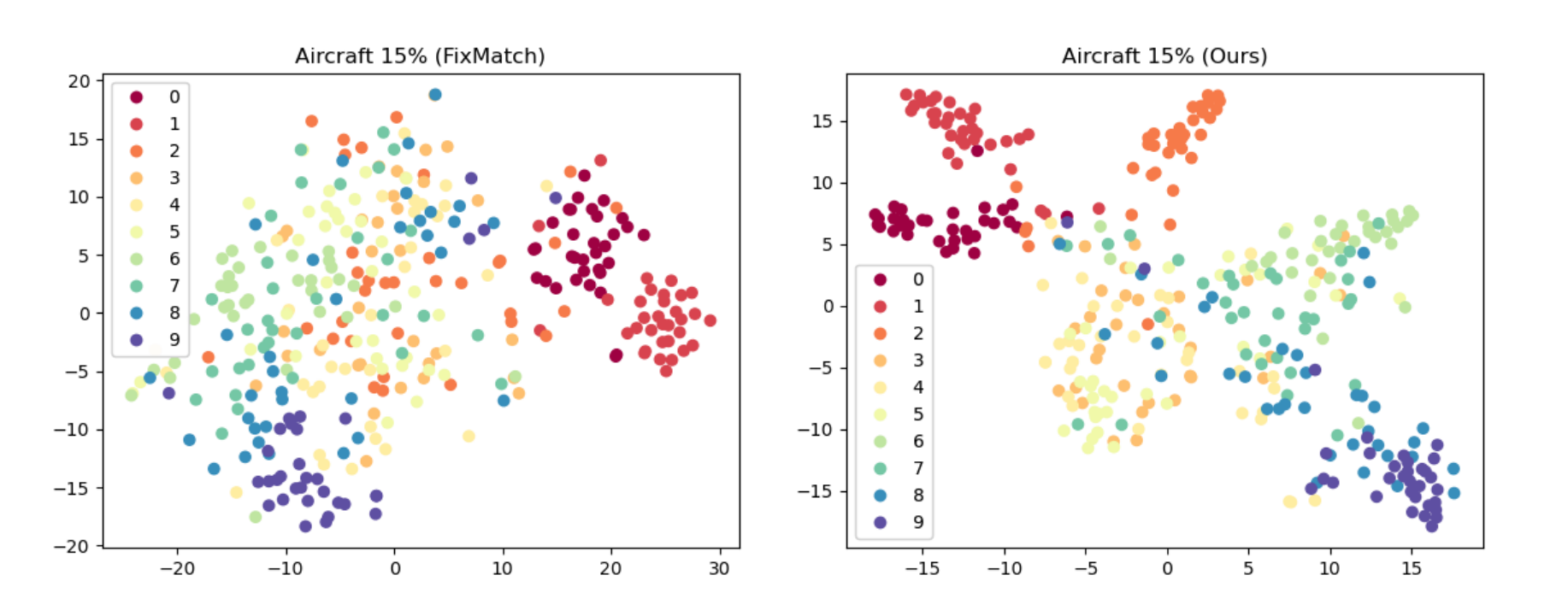}
\caption{Feature embedding visualizations of (left) FixMatch and (right) Ours approach for the first 10 classes of \textit{FGVC Aircraft} dataset by using t-SNE~\cite{van2008visualizing}. Both of the models are initialized with identical ResNet-50 supervised pre-trained on ImageNet.}
\vspace{-0.3cm}
\label{fig:tsne}
\end{figure*}

%% file: figure/fig_moco_cub.tex
\begin{figure}[t]
  \centering
  \includegraphics[width=0.5\textwidth]{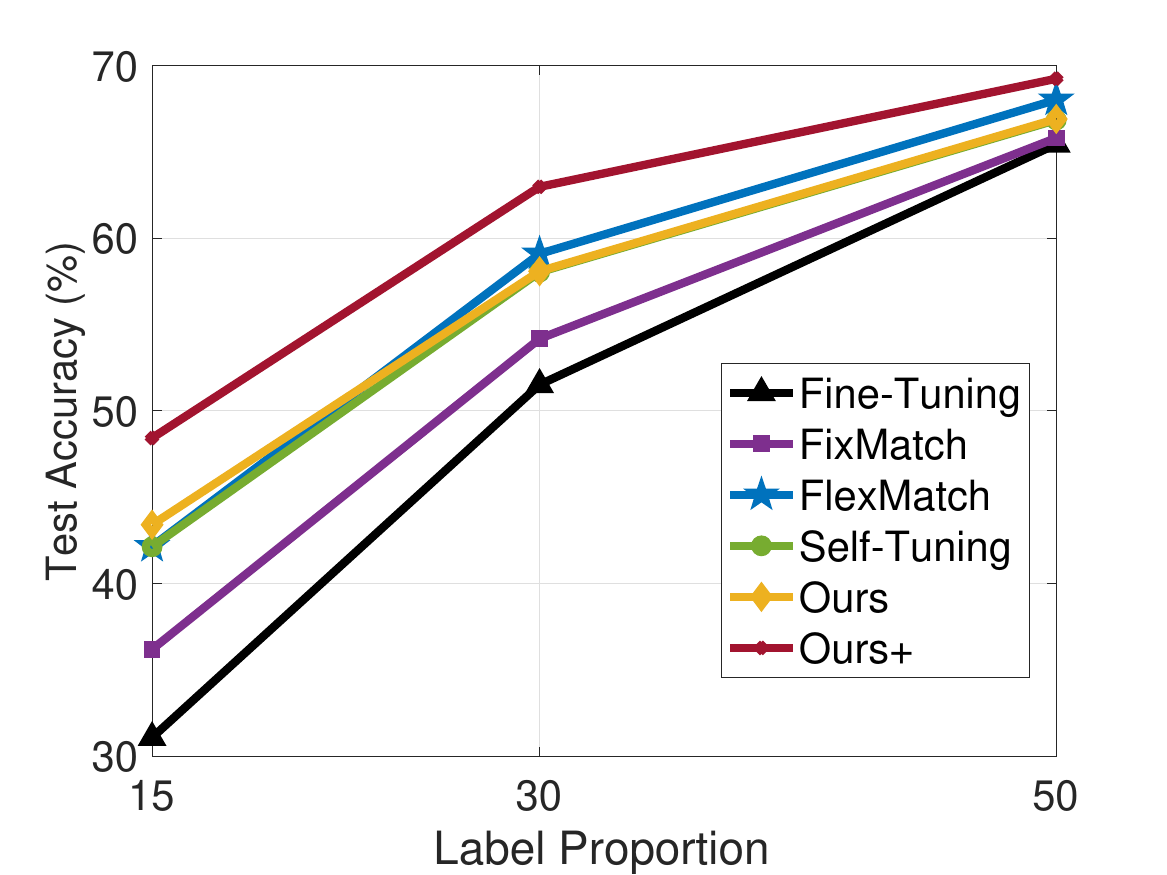}
  \caption{Test accuracy (\%) $\uparrow$ of comparing methods on \textit{CUB-200-2011} with MoCov2 which is unsupervisedly pre-trained on ImageNet1K~\cite{deng2009imagenet}.}
  \label{fig:main_res_moco_cub}
\end{figure}

%% file: table/table_supp_random_init.tex
\begin{table}[t]
    \centering
    \begin{tabular}{ l | c c c}
         \toprule
         \multirow{2}{*}{Method} & \multicolumn{3}{c}{Label Number} \\
         \cmidrule(lr){2-4}
         & 400 & 2500 & 10000 \\
         \hline
         FixMatch~\cite{sohn2020fixmatch} & 42.50 & 27.07 & 21.88 \\
         \textbf{HCCMatch (ours)} & 42.69 & 26.16 & 21.39 \\
         \bottomrule
    \end{tabular}
    \caption{Error rates (\%) $\downarrow$ on \textit{CIFAR-100} with a \textit{randomly initialized} WideResNet-28-8~\cite{ZagoruykoK16WideResNet} network. We implement HCCMatch based on the PyTorch implementation\protect\footnotemark of FixMatch which has obtained better performance than the reported ones in~\cite{sohn2020fixmatch}. 
    }
    \label{tab:supp_random_init}
\end{table}
\footnotetext{\url{https://github.com/kekmodel/FixMatch-pytorch} (CC BY)}

%% file: table/table_abl_model_bias.tex
\begin{table}[tbp]\footnotesize
    \centering
    \begin{tabular}{l | c c c}
         \toprule
         \multirow{2}{*}{\shortstack[l]{Method \\ (w/ same pre-trained model)}} & \multicolumn{3}{c}{Label Proportion} \\
         \cmidrule{2-4}
          & 15 & 30 & 50 \\
         \hline
         UDA~\cite{xie2019unsupervised} & 59.50 & 74.08 & 81.10 \\
         \textbf{UDA+Feature Adjustment(ours)} & \textbf{65.74} & \textbf{80.11} & \textbf{83.83} \\
         \bottomrule
    \end{tabular}
    \caption{Ablation study to the effectiveness of the proposed progressive feature adjustment module to the popular consistency-regularization based SSL method UDA on \textit{FGVC Aircraft}.}
    \label{tab:abl_model_bias}
    \vspace{-0.5cm}
\end{table}

%% file: figure/fig_threshold_sensitive.tex
\begin{figure*}[t]
\centering
\includegraphics[width=\linewidth]{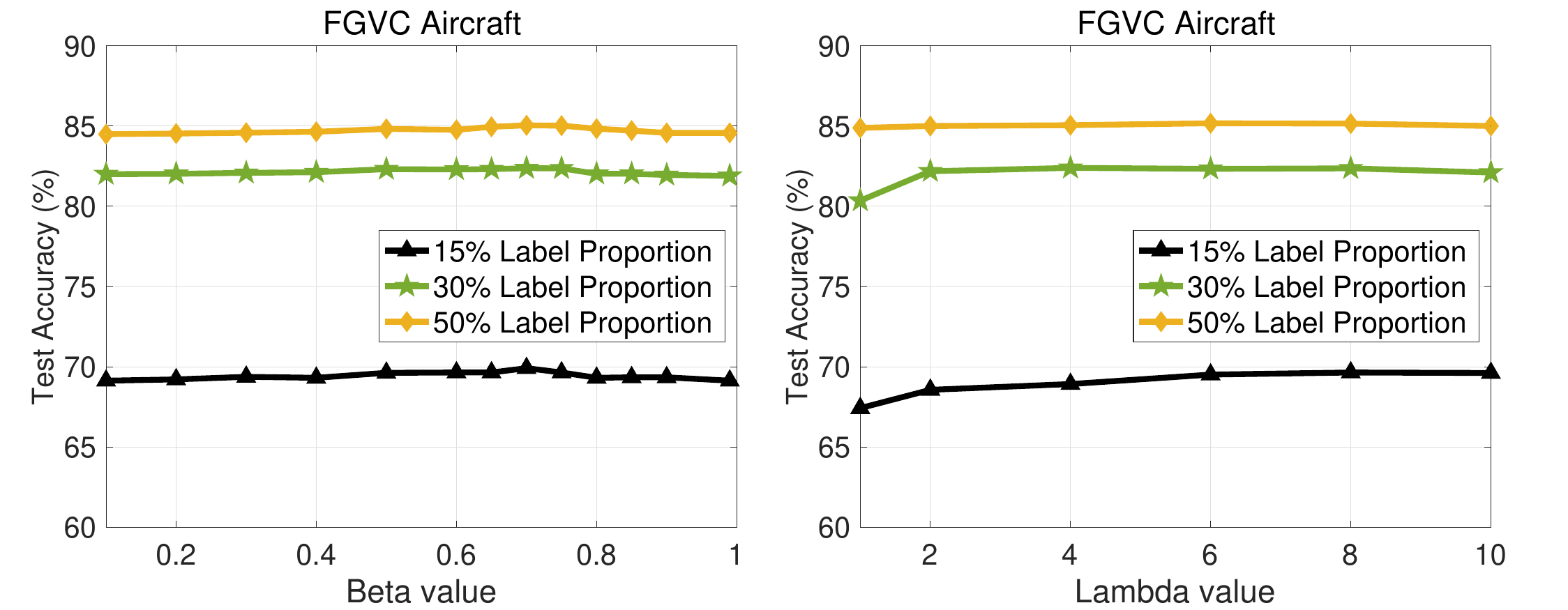}
\caption{Ablation study to the hyperparameter sensitivity.}
\vspace{-0.3cm}
\label{fig:abl_hyper_sensitivity}
\end{figure*}

%% file: table/table_abl_proto.tex
\begin{table}[t]\small
    \centering
    \begin{tabular}{l | c c c}
         \toprule
         \multirow{2}{*}{Ways of updating $\mu$ in HCCMatch} & \multicolumn{3}{c}{Label Proportion} \\
         \cmidrule(lr){2-4}
         & 15 & 30 & 50 \\
         \hline
         w/ weakly augmented images & 68.38 & 82.06 & 84.79  \\
         w/o confidence threshold & 68.73 & 81.61 &	84.49 \\
         \textbf{default (ours)} & \textbf{69.64} & \textbf{82.36} & \textbf{85.02} \\
         \bottomrule
    \end{tabular}
    \caption{Ablation study to the ways of implementing online generative classifier learning in the proposed HCCMatch approach on \textit{FGVC Aircraft} dataset.}
    \label{tab:abl_proto}
    \vspace{-0.5cm}
\end{table}